\documentclass{article}

\usepackage{PRIMEarxiv}

\usepackage[utf8]{inputenc} 
\usepackage[T1]{fontenc}    
\usepackage{hyperref}       
\usepackage{url}            
\usepackage{booktabs}       
\usepackage{amsfonts}       
\usepackage{nicefrac}       
\usepackage{microtype}      
\usepackage{lipsum}
\usepackage{fancyhdr}       
\usepackage{graphicx}       
\graphicspath{{media/}}     

\usepackage[utf8]{inputenc} 
\usepackage[T1]{fontenc}    
\usepackage{hyperref}       
\usepackage{url}            
\usepackage{booktabs}       
\usepackage{amsfonts}       
\usepackage{nicefrac}       
\usepackage{microtype}      
\usepackage{lipsum}
\usepackage{graphicx}
\usepackage{bbold}
\graphicspath{{media/}}     
\usepackage{multicol}
\usepackage{mathtools, cuted}
\usepackage{algpseudocode}
\usepackage[ruled,lined]{algorithm2e}
\usepackage{amsmath}
\usepackage[dvipsnames]{xcolor}

\title{Hierarchical Automatic Multilayer Power Plane Generation with Genetic Optimization and Multilayer Perceptron
}

\author{HAIGUANG LIAO$^1$ \hspace{12pt}
       {VINAY PATIL$^1$} \hspace{12pt}
       {XULIANG DONG$^1$} \hspace{12pt}
       {DEVIKA SHANBHAG$^1$} \hspace{12pt}
       \\ 
       {ELIAS FALLON$^2$} \hspace{12pt}
       {TAYLOR HOGAN$^2$} \hspace{12pt}
       {MIRKO SPASOJEVIC$^2$} \hspace{12pt}
       {LEVENT BURAK KARA$^1$}\thanks{Address all correspondence to this author.}\\
	1. Carnegie Mellon University\\
	Pittsburgh, PA 15213\\
	2. Cadence Design Systems\\
	San Jose, CA 95134
}

\begin{document}
\maketitle

\begin{abstract}
We present an automatic multilayer power plane generation method to accelerate the design of printed circuit boards (PCB). In PCB design, while automatic solvers have been developed to predict important indicators such as the IR-drop, power integrity, and signal integrity, the generation of the power plane itself still largely relies on laborious manual methods. Our automatic power plane generation approach is based on genetic optimization combined with a multilayer perceptron and is able to automatically generate power planes across a diverse set of problems with varying levels of difficulty. Our method GOMLP consists of an outer loop genetic optimizer (GO) and an inner loop multi-layer perceptron (MLP) that generate power planes automatically. The critical elements of our approach include contour detection, feature expansion, and a distance measure to enable island-minimizing complex power plane generation. We compare our approach to a baseline solution based on A*. The A* method consisting of a sequential island generation and merging process which can produce less than ideal solutions. Our experimental results show that on single layer power plane problems, our method outperforms A* in 71\% of the problems with varying levels of board layout difficulty.  \textcolor{black}{We further describe H-GOMLP, which extends  GOMLP to multilayer power plane  problems using hierarchical clustering and net similarities based on the Hausdorff distance.}

\end{abstract}
\section{Introduction}
\begin{figure*}[thpb]
\centering
\includegraphics[scale=.45]{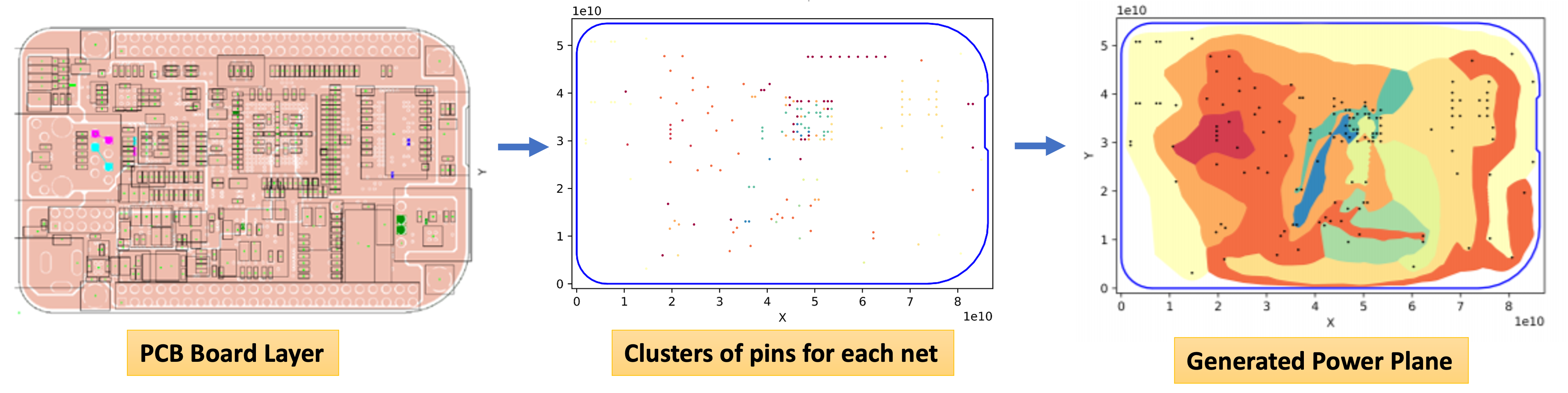}
\caption{\textcolor{black}{Single layer} power plane generation on a PCB board. Middle: Pins belonging to the same net are colored similarly. Right: Planes belonging to the same net are colored similarly. Note that some nets are split into multiple islands. }
\label{fig:mlpga-pcb-board}       
\end{figure*}

Following Moore's Law~\cite{schaller1997moore}, modern electronics designs are getting increasingly more complex. Printed Circuit Boards (PCB) design, as one of the critical  steps in electronics design is also getting increasingly more complex and time-consuming with the increasing size of boards, more complex design rule constraints (DRCs), as well as signal integrity (SI) and power integrity (PI) requirements. One of the most critical steps that hamper PCB design development is the generation of a power plane~\cite{smith2001power}, which aims to configure the various metal planes' layout on a PCB. Although there has been significant development in automatic solvers for the critical objectives of PCB design related to power plane generation such as IR-drop solver~\cite{zhong2005fast,nithin2010dynamic,zhang2015ir}, power integrity solver~\cite{wu2010overview,fizesan2010simulation}, and signal integrity~\cite{kim2010special,chen2010analysis,eudes2011experimental}, the generation of power plane itself still relies on the manual effort of electronics designers. Thus, automatic power plane generation is needed to address an unmet need in PCB design. 

In this work, we present an evolutionary automatic power plane generation method based on a multilayer perceptron (MLP) and genetic optimization (GO), which we refer to as  \textbf{GOMLP}. This  method is based on a nested combination of a multilayer perceptron and genetic optimization, with critical components including contour detection, feature expansion and customized distance metrics. Systematic experiments are conducted to study the contributions of the critical components and also stress test the method on a diverse set of problems with varying difficulties. The method is compared against a baseline solution based on A*~\cite{tseng2014star}. To the best of our knowledge, this work is the first effort to solve power plane generation with fully automated methods. \textcolor{black}{The GOMLP automatic power plane generation method is then further extended to solve multilayer power plane generation problem by integrating it with a hierarchical method that solves the challenging combinatorial problem of layer assignment. The hierarchical method applies hierarchical clustering with customized distance metrics based on Hausdorff Distance and Earth Mover Distance to generate dendrograms of nets. The generated dendrogram is applied to determine layer assignment of nets in flexible and efficient ways. We refer to this approach as \textbf{Hierarchical GOMLP} \textbf{(H-GOMLP)}.} 


\section{Background}

\subsection{Problem Formulation: Power Plane Generation}
Power plane generation is a critical step  PCB design, where the layout of metal planes connecting critical components is configured on each layer of the PCB board  \cite{cui2003dc,smith2001power}. As shown in Fig~\ref{fig:mlpga-pcb-board},  a layer of the PCB board contains a set of $m$ nets $\{N_1,N_2,N_3,...,N_m\}$. Each  net is composed of a potentially large number of pins (the I/O ports of the components on the board)
$P_{N_i} = \{p_1,..,p_{q_i}\}$, where $q_i$ is the total number of pins for $N_i$.  $p_j \in \mathbb{R}^2$ encodes the $x$ and $y$ coordinates of a pin. Net information and corresponding pin locations are determined in a previous step of board design and component placement. As such, all pin locations are fixed and are not subject to change. The objective is to group together all the pins with the same label (net assignment) through a metal plane, while excluding all other pins.

Power plane generation can be formulated as a \textbf{mutually exclusive space partitioning} of the board $\Omega$ into $m$ distinct power planes $\Omega_1, ..,\Omega_m$ such that:
\begin{equation}
    \Omega = \Omega_1 \cup \Omega_2 \cup ... \cup \Omega_m ~~and ~~ \Omega_i \cap \Omega_j = \emptyset
\label{space-partition-define}
\end{equation}
Plane $\Omega_i$ corresponds to net $N_i$, which means the number of planes is equivalent to the number of nets. Each plane $\Omega_i$ may  consists of several \textbf{islands}: $\Omega_i =  I_i^1 \cup I_i^2 ... \cup I_i^{s_i}$, where $s_i$ is the total number of islands of plane (or net) $i$. 

\noindent \textbf{Design Constraint}: All  pins of Net $N_i$ must be located inside $\Omega_i$ (possibly spread over multiple islands) and no pin from another net can appear in $\Omega_i$:

\begin{equation}
\label{pin-inclusion-constraint}
P_{N_i} = \{p_1,..,p_{q_i}\} \in \Omega_i
\end{equation}
 
 \noindent \textbf{Design Objective}: The power plane generation problem formulation described above is the basic form. For each plane $\Omega_i$, it is preferable to generate a singly connected plane consisting of a single island. Otherwise, disconnected islands belonging to the same plane will have to be connected on a separate layer of the PCB using extraneous connections and vias. Mathematically, this can be stated as finding an optimized space partitioning $\Omega^*$ which minimizes the total number of split islands:
 
 \begin{equation}
\label{base_opt_form}
\Omega^* = argmin_{\Omega} \sum_{i=1}^m s_i
\end{equation}
 
Given a board, we denote a set of power planes that satisfy the design constraint  as \textbf{feasible}, and for each net, those that have a minimum number of islands as \textbf{desirable}. Fig.~\ref{fig:mlpga-demovalidPP} shows examples.  For the example design, there are two nets $\{N_1,N_2\}$, where $N_1$ has two pins and $N_2$ has one pin. The desirable designs (E and F) are with singly connected partitions  that are feasible. Designs C and D, while feasible, split one of the nets into two separate islands. Compared to E and F, these designs are not desirable. 

The objective of minimizing the number of islands for each net makes the problem non-trivial since planes corresponding to different nets have to compete for the shared design space while satisfying the design constraint. Intuitively, \textcolor{black}{single layer} power plane generation is analogous to a 2D multi-class image segmentation problem  aimed at minimizing the number of islands for all nets while ensuring that no pin is misclassified. While approaches such as image segmentation~\cite{cheng2001color,pham2000current,haralick1985image} and space partitioning~\cite{wang2020learning,buck1943partition} have been extensively studied, these approaches do not put emphasis on split island minimization. To the best of our knowledge the power plane generation is an open problem not only in the scope of PCB design but also in the broader context of space partitioning problems.   

\begin{figure*}[thpb]
\centering
\includegraphics[scale=.38]{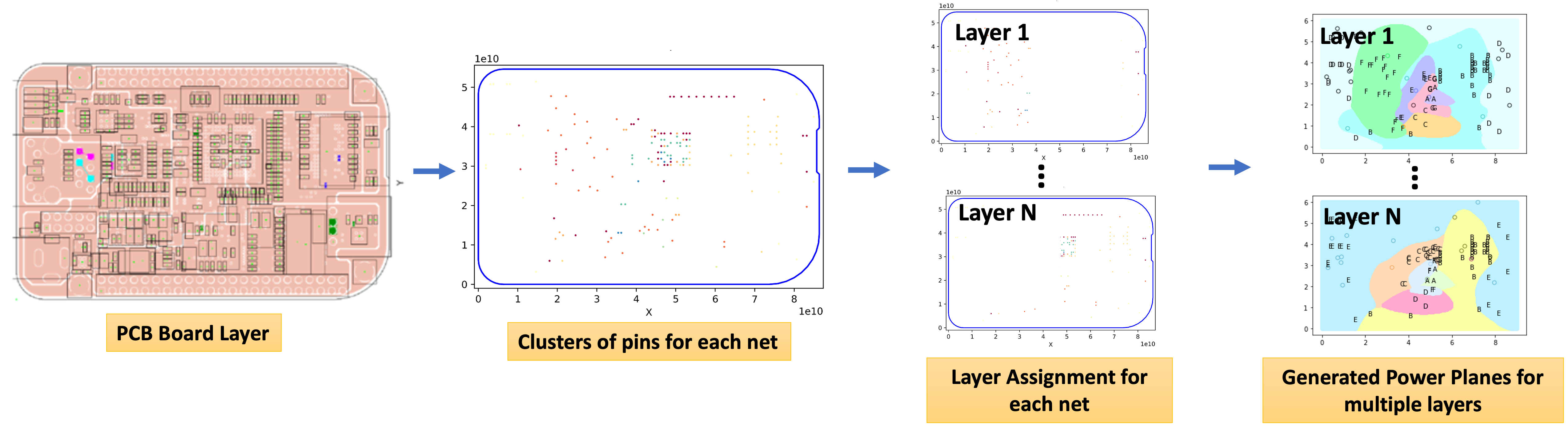}
\caption{\textcolor{black}{Multilayer power plane generation on a PCB board.}}
\label{fig:multi-pcb-board}       
\end{figure*}

\textcolor{black}{\noindent \textbf{Multilayer Formulation}: Most industry PCB designs consist of multiple layers of boards and thus power plane generation solutions need to be extended to account for multiple layers. Fig.~\ref{fig:multi-pcb-board} shows our workflow for multilayer cases. Compared to the single layer case as shown in Fig.~\ref{fig:mlpga-pcb-board}, the multilayer case deals  with one additional \textbf{layer assignment} step, where given a prescribed number of layers $K$, each net in a design $N_i$ is assigned to a unique layer $L_i$. After  layer assignment, each layer $L_i$  consists of a set of $z$ nets: $\{N_{L_i1},N_{L_i2},...,N_{L_iz_{L_i}}\}$. At this point, the problem becomes a set of single layer  problems where the power planes for each layer are generated individually. The layer assignment step is challenging owing to its combinatorial nature: Enumerating all possible choices of layer assignment with $N$ nets and $K$ layers requires solving single layer power plane problems $N^{K+1}$ times, which is intractable with limited computation resources. This motivates us to develop  an efficient and flexible hierarchical method to solve multilayer power plane generation problems.
}

\subsection{Scope}

\begin{figure}
\centering
\includegraphics[scale=.52]{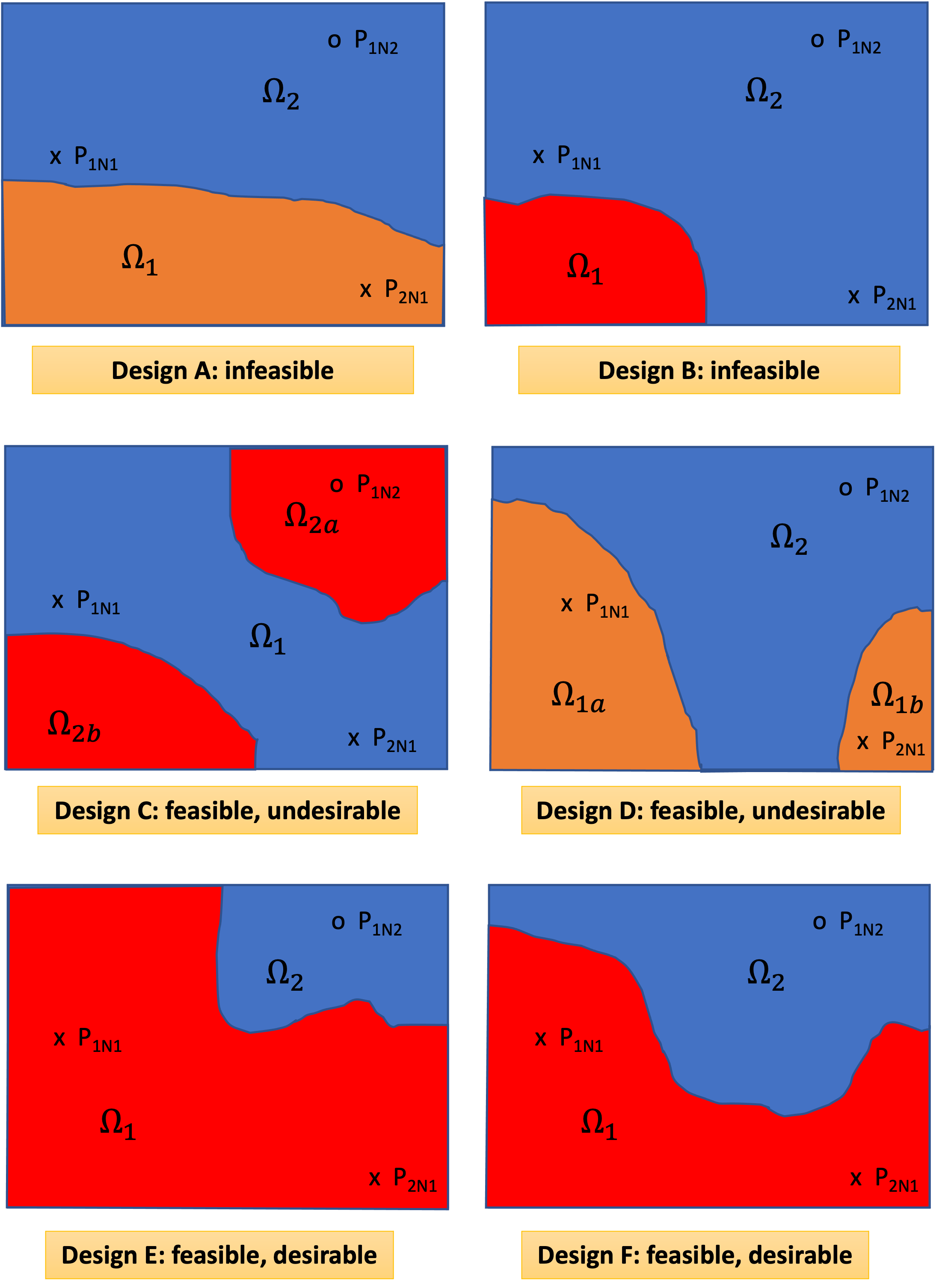}
\caption{Plane feasibility vs. desirability. }
\label{fig:mlpga-demovalidPP}       
\end{figure}

Our approach seeks to find a feasible solution that minimizes the number of islands. In this work, the priority is on identifying power planes that match this desire. However, we do not optimize the  outer contour geometries of the planes and reserve this consideration for future work. As such, in the presented work, designs E and F in Fig.~\ref{fig:mlpga-demovalidPP} are identical in quality.

In industrial PCB design, there exists other constraints and objectives for  space partitioning. These include considerations around  IR-drop, power integrity, and signal integrity of the partitioned board. For instance, it is common to impose an upper bound on the admissible  IR-drop  for each source-drain pin pair within a plane, which favors planes with the least number of thin regions. For each power plane candidate, the evaluation of the IR drop involves a  physical simulation that can be prohibitive. In this work, we thus exclude such considerations in favor of establishing the foundations for island minimizing partitioning. Future work will have to account for such criteria. 

\textcolor{black}{In this work, GOMLP and H-GOMLP are used to solve different power plane  problems. GOMLP only considers a single layer of PCB, aiming to identify an optimized solution to the nets and pins already assigned to a specific layer (i.e., net and pin assignment is assumed to have been completed). By contrast, H-GOMLP, which  extends  GOMLP, focuses on solving multilayer power plane generation. In general, while  GOMLP can be viewed as a component of H-GOMLP, it can  be utilized as a standalone,  single layer PCB  generation tool.}

\section{Methods}

\begin{figure}
\centering
\includegraphics[scale=.45]{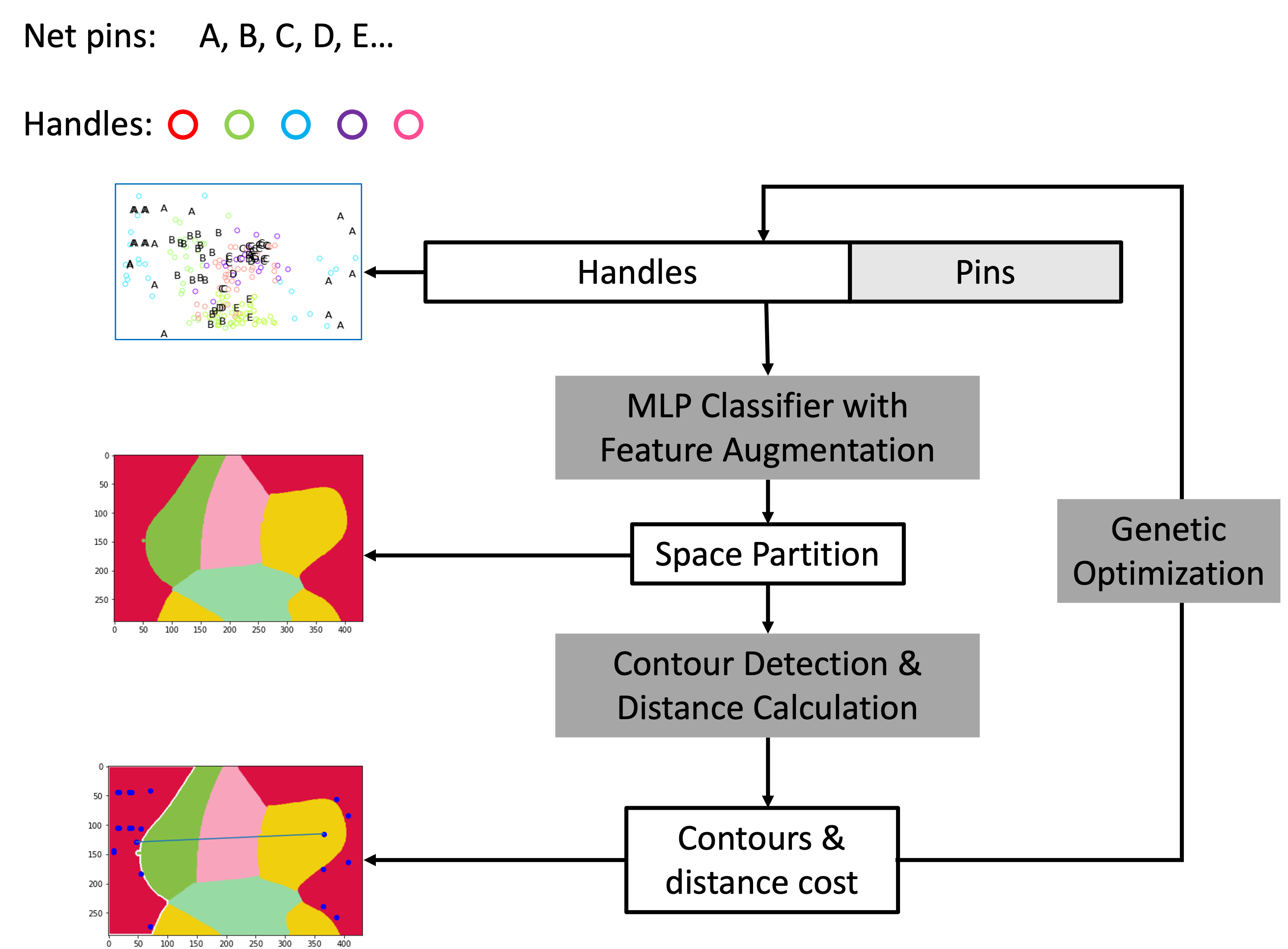}
\caption{\textcolor{black}{Proposed GOMLP method to solve single layer power plane generation problems.}}
\label{fig:mlpga-mlpga-flow}       
\end{figure}

\textcolor{black}{Since GOMLP is an independent module but also a component of the more complex H-GOMLP, we will first introduce GOMLP as an standalone method and then introduce H-GOMLP.}

\textcolor{black}{In GOMLP}, we parameterize the problem using a set of \textbf{handles} defined for each plane $\Omega_i$ as $H_{N_i} = \{h_1,..,h_{k_i}\}$, where $h_j \in \mathbb{R}^2$ encodes the $x$ and $y$ coordinates of the handle on the board. These handles serve as labelled movable points on the board such that an optimized placement of the entire set of handles $H_{N_i}, i:1..m$ produce a partitioning of the board where all original (and static)  pin sets $P_{N_i}, i:1..m$ are correctly grouped, and the number of power plane islands is minimized. 


\textcolor{black}{GOMLP is} summarized in Fig.~\ref{fig:mlpga-mlpga-flow}. It consists of an outer loop genetic optimizer (GO) that identifies the best handle placements. To evaluate a candidate solution, we use an inner loop multi-layer perceptron (MLP) that produces a 2D segmentation of the board using $H_{N_i}$ and $P_{N_i}, i:1..m$ as the labelled data to be classified. The MLP uses an expanded feature space and, is allowed to overfit to the data so as to ensure all pins are correctly classified (i.e., satisfying the design constraint). A separate module then identifies the total number of islands together with the pairwise distances between the islands of the same net. These quantities are appended to GO's fitness function to drive the optimal handle placement.

\subsection{Genetic optimization for handle placement}

Genetic Optimization (GO) has found extensive use in problems where the gradients of the objective function and constraint are not readily available  ~\cite{whitley1994genetic,jing2003parallel,muhlenbein1992parallel}.  In this work, we use GO to optimize the handles' locations, which then are used as input to the MLP to produce the space partitions . The primary motivation for applying GO instead of other gradient-based optimizers is that the objective function in the power plane generation problem is not readily differentiable.  

In our GO, a population consists of $M$ chromosomes, where each chromosome encodes the handle coordinates as a vector of genes. Thus, each chromosome is a candidate solution for $\Omega$. Each gene has two scalars $x,y$ encoding the 2D coordinates of a handle.  For each $\Omega_i, i:1..m$, we instantiate $k$ handles, where k is calculated as follows:

 \begin{equation}
\label{kcalc}
k = \frac{2 \cdot \sum_{i=1}^m q_i}{m}
\end{equation}

\noindent where $q_i$ is the total number of pins for net $i$. Each chromosome is thus a $2\cdot k\cdot m \times 1$ real vector, with the handles belonging to each net occupying a $2\cdot k \times 1$ block. Handles are randomly initialized in the first generation of GO. We use a fitness-proportional parent selection, uniform mutation and probabilistic random swap for crossover operators \cite{LIPOWSKI20122193,GAmutation}. Fig.~\ref{fig:mlpga-mlpga-flow}-top shows the collection of pins (letters) and handles (circles) for a 5-net design problem. 

\subsection{Power plane generation using MLP}

Given a chromosome, we use an MLP to construct the partitioned power planes. The MLP takes as input the labelled training set  $P_{N_i} \cup H_{N_i}, i:1..m$ and  creates a region segmentation of the board using the multi-class cross entropy loss. Fig.~\ref{fig:mlpga-mlpga-flow}-middle shows the current space partitioning result. For each item in the training set, we expand its $x,y$ coordinates into a $15 \times 1$ vector: $\{ x,y,x\cdot y, \sin(2\pi\cdot x),\cos(2\pi\cdot x),x^2,\sin(2\pi\cdot y),\cos(2\pi\cdot y),y^2,\sin(3\pi\cdot x),\cos(3\pi\cdot x),x^3,\sin(3\pi\cdot y),\cos(3\pi\cdot y),y^3\}$ to enable complex boundaries. The MLP is trained with these features as input and the net classification as output is then used to generate space partition $\Omega$ by predicting the plane assignment for every coordinate from the meshgrid on the PCB board layer. 

The MLP consists of three hidden layers, with  50  neurons in each layer and the $tanh$  activation function. Adam optimizer with a learning rate of 0.002 is used. It is desirable to overfit to the training data,  so as to favor all pins to satisfy the feasibility constraint of Eqn. \ref{pin-inclusion-constraint}. Nonetheless, the resulting partitioning may result in disconnected islands, and may not satisfy Eqn. \ref{pin-inclusion-constraint}. 

\subsection{Fitness calculation}

To evaluate each solution's fitness, we determine the total number of islands $s_i$ of each plane $\Omega_i$ generated by the MLP using a contour detection algorithm \cite{contourdetection}.

\begin{equation}
F^{island} =  -\sum_{i=1}^m s_i
\end{equation}

\noindent Additionally, we append the fitness function with a measure to distinguish the islands of a plane $\Omega_i$ that are far apart from those that are proximate. This helps GO acquire guidance in handle placement for the subsequent generation (Fig.~\ref{fig:mlpga-mlpga-flow}-bottom). For $\Omega_i$, we define:

\begin{equation}
F_i^{d_{min}} =  -\sum_{a,b}^{s_i} ||I_i^a - I_i^b||_{min}
\end{equation}

\noindent where $I_i^a, I_i^b$ denote islands $a $ and $b$ belonging to $\Omega_i$, and $||.|| $ is the Euclidean norm. $F_i^{d_{min}}$ captures the sum of minimum distances between the pairs of the disjoint islands of $\Omega_i$.

While min. distances are useful in signaling nearly connected islands, they typically favor the generation of elongated thin connections. To additionally drive the islands toward one another, we define a centroid distance as a part of the fitness function:

\begin{equation}
F_i^{d_{cent}} =  -\sum_{a,b}^{s_i} ||I_i^a - I_i^b||_{cent}
\end{equation}

\begin{figure}
\centering
\includegraphics[scale=.39]{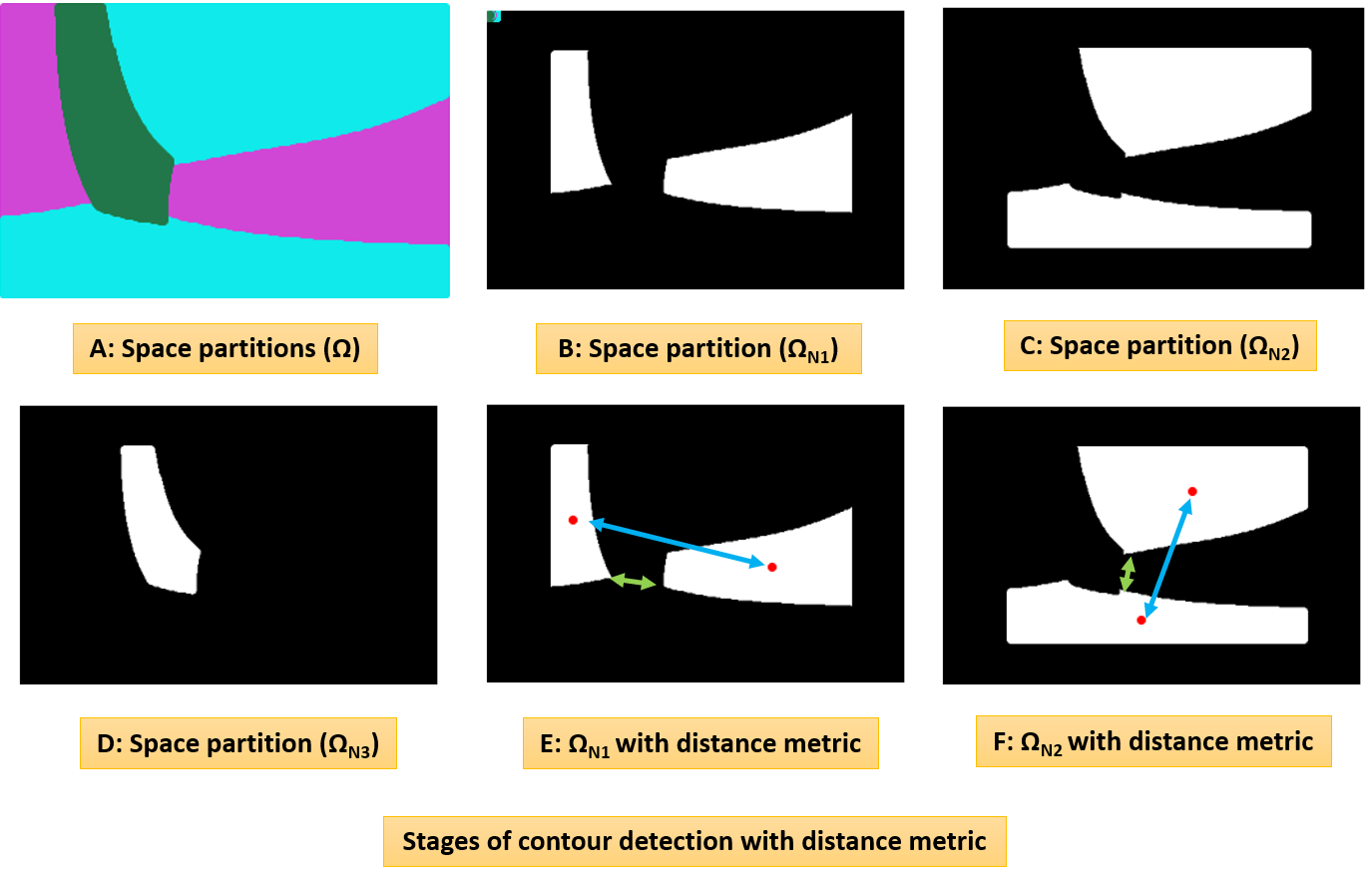}
\caption{Contour detection and distance metric calculations of space partitions.}
\label{fig:contour-detection}       
\end{figure}

\noindent Fig.~\ref{fig:contour-detection} illustrates the above two islands distance calculations. For a segmentation solution  produced by the MLP, the final fitness maximized by the GO is:

\begin{equation}
F = F^{island} +  \sum_{i}^{m} F_i^{d_{min}} +  \sum_{i}^{m} F_i^{d_{cent}}
\end{equation}

\begin{figure}[thpb]
\centering
\includegraphics[scale=.45]{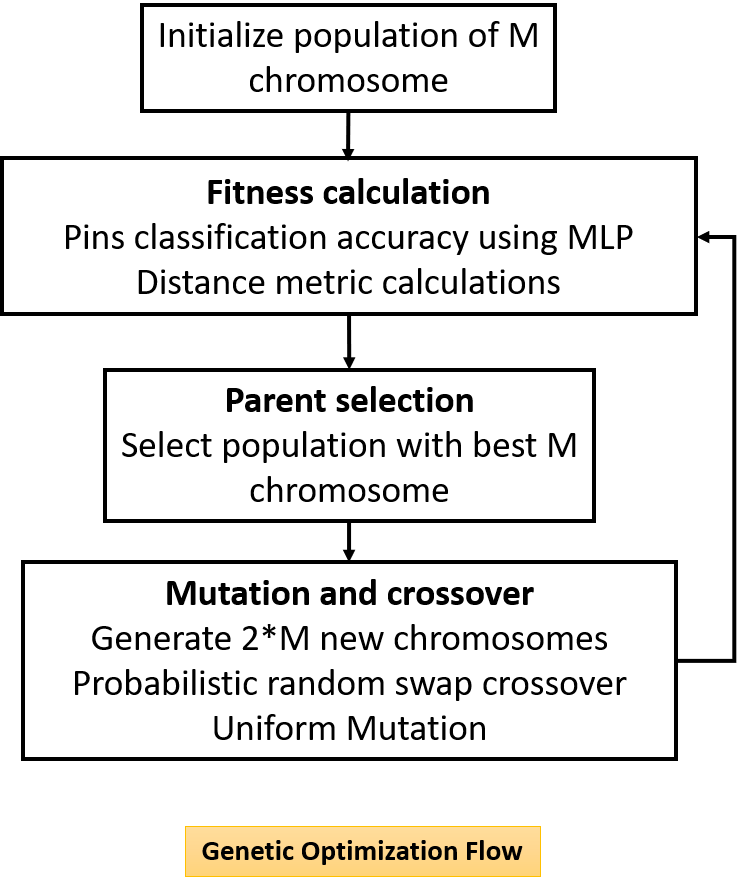}
\caption{Details of GO structures and flow}
\label{fig:mlpga-flow}       
\end{figure}


\begin{algorithm}[ht]
\SetKwInOut{Input}{Input}\SetKwInOut{Output}{Output}
\Input{Netlist of designs: $\{N_1,N_2,...,N_m\}$}
\Output{Space partitions: $\{\Omega_1,\Omega_2,...,\Omega_m\}$}
Initialize handles $\{h_{1},h_{2},...,h_{k_i}\}$ for each net $N_i$ as the first population for GO\;
\For{generation=1,...,$N$}{
MLP fitting all fixed points $\{P_{1},P_{2},...,P_{q_i}\}$ and handles  $\{H_{1},H_{2},...,H_{k_i}\}$ for each net $N_i$ with augmented features\;
Contour detection and distance metric calculations for all planes $\{\Omega_1,\Omega_2,...,\Omega_m\}$ \;
Calculating fitness scores for all chromosomes based on MLP fitting results\;
Generating new population with crossover and mutation based on selected elites\;
}
 \caption{GOMLP power plane generation}
 \label{mlp-ga-code}
\end{algorithm}

\subsection{\textcolor{black}{H-GOMLP}}
\begin{figure*}[thpb]
\centering
\includegraphics[scale=.47]{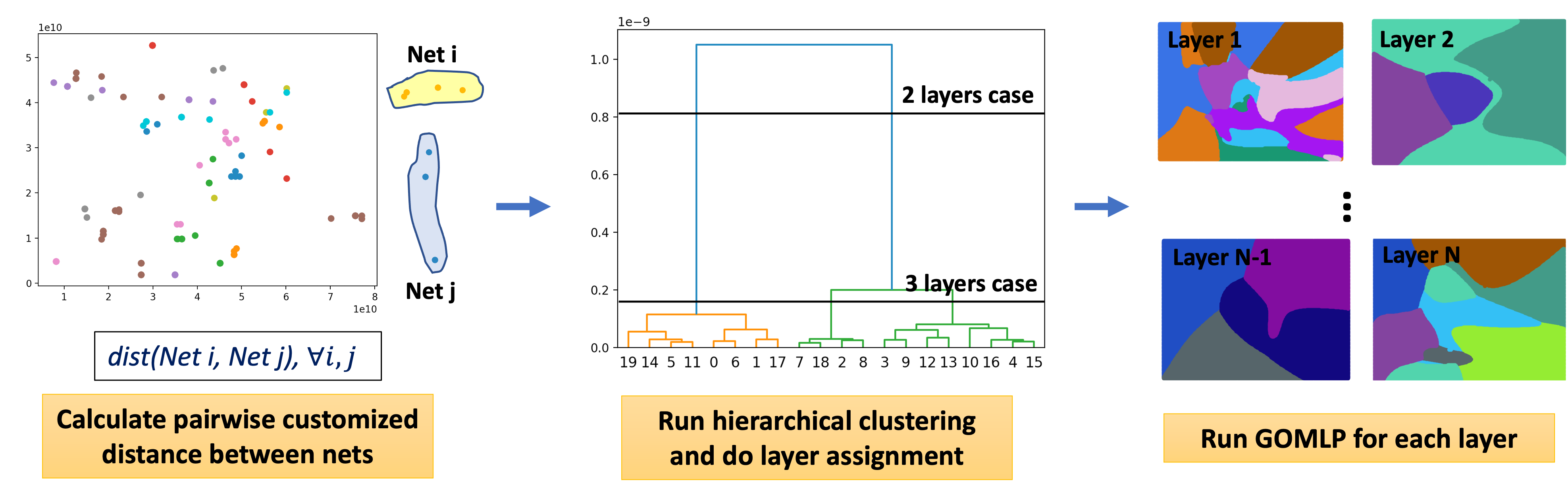}
\caption{\textcolor{black}{Proposed H-GOMLP method to solve multilayer power plane generation problems.}}
\label{fig:h-gomlp-flow}       
\end{figure*}

\textcolor{black}{H-GOMLP extends  GOMLP to  multilayer power plane generation problems. Fig.~\ref{fig:h-gomlp-flow} shows the flowchart of our proposed H-GOMLP method, which  consists of 3 major steps:
\begin{itemize}
    \item 1. Calculating pairwise  distances between all net pairs 
    \item 2. Running hierarchical clustering for all nets to facilitate layer assignment
    \item 3. Running GOMLP on each layer
\end{itemize}
}

\textcolor{black}{In the first step, given all nets $\{N_1,N_2,...,N_m\}$ in a multilayer PCB board to be designed. We calculate the \textbf{Hausdorff Distance (HD)}~\cite{huttenlocher1993comparing} and \textbf{Earth Mover Distance (EMD)}~\cite{andoni2008earth} between all pairs of nets. For a pair of nets $\{N_i,N_j\}$, the distance is calculated based on the coordinates of their corresponding pins. For HD, the distance is calculated by:}
\begin{equation}
\begin{split}
d_{HD}(N_1,N_2) = max\{max_{x_i \in N_1} min_{y_j \in N_2}||x_i-y_j||,\\max_{y_j \in N_2} min_{x_i \in N_1}||x_i-y_j||\}
\end{split}
\end{equation}
\textcolor{black}{For EMD, the distance is calculated by:}
\begin{equation}
\begin{split}
d_{EMD}(N_1,N_2) = min_{\phi:N_1 \rightarrow N_2} \sum_{(x,y) \in N_1} ||(x,y)-\phi((x,y))||_2
\end{split}
\end{equation}
\textcolor{black}{where $\phi: N_1 \rightarrow N_2$ is a bijection. The motivation for using the HD and EMD to calculate the distance between each pair of nets is to approximate the proximity and overlapping of the nets' layouts. Smaller HD and EMD distances indicate that the pins' layouts of two nets are very close to each oother or significantly overlap. In this way, we can make use of this information in the layer assignment step as an effective metric for the spacing of nets.}

\textcolor{black}{After the distance calculation, a \textbf{hierarchical clustering} for all the nets is done based on the pairwise distance of all nets. In the hierarchical clustering, instead of using the typical distance metrics like Euclidean distance, we use the inverse of the HD and EMD between pairs of nets calculated in previous step. The motivation of using inverse of HD and EMD is to cluster nets that are less overlapping on to the same layer in order to use each layer most efficiently. In such case, the relative difficulty of finishing up those single layers with GOMLP would be minimum. In another respect, the overall usage of each layer space is more efficient.} 

\textcolor{black}{The hierarchical clustering will generate a dendrogram of nets as shown in Fig.~\ref{fig:h-gomlp-flow}. Based on this dendrogram, we can generate clusters of nets with specified budget of cluster numbers. Nets belonging to the same cluster will then be assigned to a unique layer of the PCB board. As shown in Fig.~\ref{fig:h-gomlp-flow}, based on the dendrogram, we can get a 2-layer case at a higher level on the dendrogram and get another 3-layer case at a slightly lower level. This is a unique advantage of our approach, since it provides flexibility in the number of layers needed for a design: for a given multilayer PCB board, the number of layers needed to finish power plane generation is usually not given upfront. With our approach, the designer can start at a very high level on the dendrogram, which gives a small layer number solution. If the small layer number does not lead to feasible or desirable designs, the designer can move down on the dendrogram to get another design with incrementally more number of layers, and the process can be repeated until desirable or feasible power planes are generated. }  

\textcolor{black}{Finally, after the layer assignment, the H-GOMLP is finished by running GOMLP for all the layers separately and the details of GOMLP have been described above. This step and the previous layer assignment step can be executed iteratively if necessary. For instance, after the first run of layer assignment with total number of layers determined to be $K$, GOMLP is applied to solve those $K$ layers independently. If it turns out that some layers cannot achieve desirable or feasible designs with GOMLP, we can go back to the layer assignment step and generate a new assignment with $K+1$ layers, followed by running GOMLP for these $K+1$ layers. One great advantage for our approach is that GOMLP execution for different layers can run in parallel as well as the GO module inside GOMLP loop, which is already realized in our implementation. This enables our method's scalability to large scale industry PCB board designs. Detailed description of H-GOMLP is available in Algo.~\ref{h-gomlp-code}.}


\begin{algorithm}[ht]
\SetKwInOut{Input}{Input}\SetKwInOut{Output}{Output}
\Input{Netlist of designs: $\{N_1,N_2,...,N_m\}$}
\Output{Space partitions for $K$ layers: $\{\Omega_{i1},\Omega_{i2},...,\Omega_{in}\}$, where $i=1,...,K$}
Calculate the HD and EMD for all pairs of nets $\{N_i,N_j\}$\;
Hierarchical clustering and generate dendrogram for all nets based on the inverse of HD/EMD\;
Layer assignment for all nets with specific number of layers $K$ based on the dendrogram\;
\For{layer=1,...,$K$}
{
Run GOMLP for nets assigned on the layer to generate space partitions: $\{\Omega_{i1},\Omega_{i2},...,\Omega_{in}\}$
}
\caption{\textcolor{black}{H-GOMLP power plane generation}}
 \label{h-gomlp-code}
\end{algorithm}

\section{Method: A* Baseline Method}
\begin{figure}
\centering
\includegraphics[scale=.48]{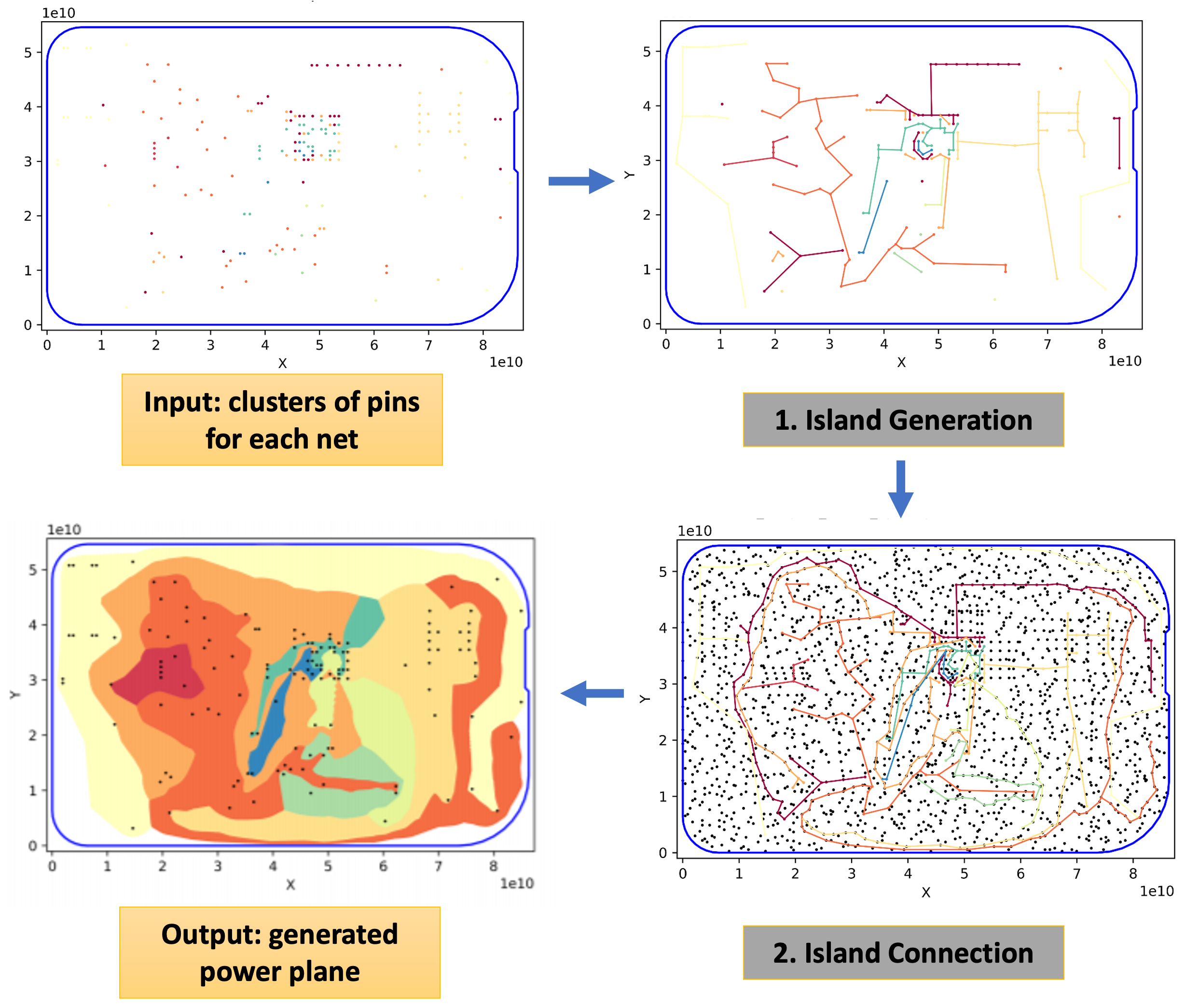}
\caption{Flowchart of A* Power Plane Generation Method.}
\label{fig:mlpga-astar-flow}       
\end{figure}

A* is one of the most frequently used search algorithms in graph traversal and path finding. In this work, A* search is used as the backbone of our baseline power plane generation algorithm, which we refer to as \textbf{A*}. It has been proven that with admissible heuristics, the A* can find the shortest path between two vertices in a graph~\cite{gelperin1977optimality}. Fig.~\ref{fig:mlpga-astar-flow} shows the flowchart of the A* baseline. The reason for selecting the A* as our baseline is that it has been the most frequently used algorithm practice to solve routing problem~\cite{hu2001survey,ozdal2007archer,liao2020deep}. In the baseline method, the solution consists of multiple steps and one of the key steps (Island Connection) can be formulated as a routing problem. 

The input of the power plane generation problem is clusters of pins belonging to each net $\{N_1,N_2,...,N_m\}$. For A*, the first step is the \textbf{Island Generation}. At this step, for each net $N_i$, a minimum spanning tree ($MST_{N_i}$) among all the pins $P_{N_i}$ belonging to net $N_i$ is generated with using Kruskal's algorithm~\cite{kruskal1956shortest}. After the MST for all nets $\{MST_{N_1},MST_{N_2},...,MST_{N_m}\}$ are generated, edge pruning is applied to all the MSTs where any edges of the MSTs that intercept edges of other nets are pruned. After pruning, there is a set of disconnected trees (pins connected with edges) for each net, which is represented by an island that includes all the pins belonging to the tree while excluding all other pins not belonging to the tree. For instance, if island generation produces $s_i$ disconnected trees $\{T_{1},T_{2},...,T_{s_i}\}$ for Net $N_i$, then the disconnected trees are represented by a corresponding set of islands $\{I^{1}_i,I^{2}_i,...,I^{s_i}_i\}$, where the area of $I_{j}$ include all the pins of tree $T_{j}$ but not include pins belonging to other trees of net $N_i$ or other nets. Based on the Island Generation step, each net is represented as a set of islands. The islands for all the nets are fed into the downstream \textbf{Island Connection} step, where A* algorithm is applied to route all the disconnected islands for each net to generate a connected tree for each net. Specifically, given an order of the nets, for each net $N_i$ with disconnected island set $\{I^{1}_i,I^{2}_i,...,I^{s_i}_i\}$, A* works on connecting the disconnected island set. For the disconnected island set with $s$ islands, $(s-1)$ pairs of islands are generated, and then A* will sequentially connect all the pairs of islands by searching through a graph consisting of upsampling nodes distributed on the board. The Island Connection is finished when all the island pairs belonging to all the nets have been attempted to be connected with A*. 

Since all the island pairs of one design have to share the same space when connected with A*, it is not guaranteed that all of them can be connected. If the Island Connection step is successful, then a set of trees $\{T_{N_1},T_{N_2},...,T_{Nm}\}$ will be generated where $T_{N_i}$ connected all the pins belonging to net $N_i$. If some islands are not fully connected, some nets will end up having more than one tree. Based on the generated tree set from the Island Connection step, the final power plane is generated by inflating the tree edges of each tree. Specifically, densely located nodes are sampled on all edges of the trees so that each tree $T_{N_i}$ can be represented as a set of nodes. Then the space partition is generated by running K-nearest Neighbors (KNN)~\cite{peterson2009k} algorithms based on the nodes from all trees with the number of neighbor $k$ set as 1.    

The A* baseline approach provides a practical solution flow to generate a power plane. However, a downside of A* is its greedy manner~\cite{wayahdi2021greedy,dogar2014object}: it focuses on finding out the optimal solution for only one search problem. In power plane generation, there are cases where multiple search problems that need to be solved on the same graph sharing limited graph space. Under such case, A* would struggle to consider all the problems simultaneously. Thus, there are at least the following combinatorial optimization challenges to be addressed to make it an applicable approach: 
\begin{itemize}
\label{co-challenge-A*}
    \item \textbf{Sequence of nets}: in the Island Connection step, the A* has to be applied to  the nets sequentially. Thus, an optimal sequence of nets needs to be determined. Otherwise, the method will be unfavorable to the nets that are routed later because the nets routed earlier have occupied part of the search graph. 
    \item \textbf{Sequence of island pairs}: similar to the sequence of nets challenge, within each net, the island pairs also need to be sequenced before A* is applied. The sequence is a non-trivial combinatorial optimization problem. 
    \item \textbf{Design of upsampling nodes}: to run the A* algorithm efficiently, the search graph size need to be reasonably small,  and thus designing the distribution of upsampling nodes is another combinatorial optimization challenge that needs to be solved. 
\end{itemize}

\begin{algorithm}[ht]
\SetKwInOut{Input}{Input}\SetKwInOut{Output}{Output}
\Input{Netlist of designs: $\{N_1,N_2,...,N_m\}$}
\Output{Space partitions: $\{\Omega_1,\Omega_2,...,\Omega_m\}$}
Tree generatoin $\{MST_{N_1},MST_{N_2},...,MST_{N_m}\}$ with MST for each net\;
Island generation by tree pruning for each net $N_i$: $\{T_{1},T_{2},...,T_{s_i}\}$\;
for net 
\For{i=1,...,$m$}{
Running A* to connect the isolated islands $\{I_{i}^1,I_{i}^2,..,I_{i}^{s_i}\}$ for net $N_i$, forming complete tree edges for each net
}
Generating final space partitions by running KNN on the upsampled nodes from edges of different net
 \caption{A* power plane generation}
 \label{astar-code}
\end{algorithm}

\section{Evaluation}

To systematically study the performance of our approach, extensive experiments are performed, which can be divided into two parts. In part one, GOMLP is applied to solve problems with and without the key components to show the effect of the key features of our method, which include the following:

\noindent\textbf{Effect of GO on MLP}: Given the same design problem, power planes are generated with standalone MLP and MLP with GO to investigate the effect of GO upon MLP.

\noindent\textbf{Effect of feature expansion on  GOMLP }: Given the same design problem, GOMLP is applied with and without expanded features to investigate the effect of feature expansion.

\noindent\textbf{Effect of distance metric on the GOMLP}:  Given the same design problem, GOMLP is applied with and without the distance metric to investigate its effects on performance of the GOMLP.

In the second part of the experiments, we apply  GOMLP as well as the baseline A* method to a set of power plane generation problems with different configurations and difficulties to systematically study the performance of our method. 

\subsection{Dataset}
To generate the problems, given a layer of design on a PCB board, we vary the number of nets from 6 to 8, with more nets corresponding to more challenging problems. For each net number, combinations of nets are selected to generate distinct problems. With the variations, there are altogether 129 different problems with varying difficulties. The GOMLP as well as the baseline A* are applied to each of the problems with maximum run time for solving each problem set as 30 minutes for a fair comparison.

 Cross entropy is used as the loss function. For the GO, the maximum number of generations is set as 20, population size is 30 and elite size is 10. An industry real PCB board design \textbf{Beaglebone} is used, and different design problems in the experiment come from different design problems on different layers. Since the objective in this work is to minimizes the total number of split islands, in evaluating the quality of generated power planes, the primary metric used is \textbf{Extra Islands (EI)}, which is defined as follows:
\begin{equation}
    EI = \sum_{i=1}^m s_i - m
\end{equation}

\textcolor{black}{For evaluating H-GOMLP method on multilayer PCB board designs, we generate more challenging problems by increasing the number of nets to 20, which is also based on industry real design \textbf{Beaglebone}. We set the parameters of GOMLP inside H-GOMLP the same as the stand-alone GOMLP. In order to compare effectiveness of different settings of H-GOMLP, the primary metric we use is the minimum number of layers we need to finish a multilayer power plane generation problem where all layers have 0 EIs. We call this metric: \textbf{Minimum Close Design Layer (MCDL)}. It is worth mentioning that the MCDL would change based on different parameter settings of GOMLP or difficulty level of the designs and thus the conclusion based on MCDL only holds for specific problems and settings.}


\section{Results and Discussions}

\subsection{Comparison between MLP and GOMLP}
\begin{figure}
\centering
\includegraphics[scale=.4]{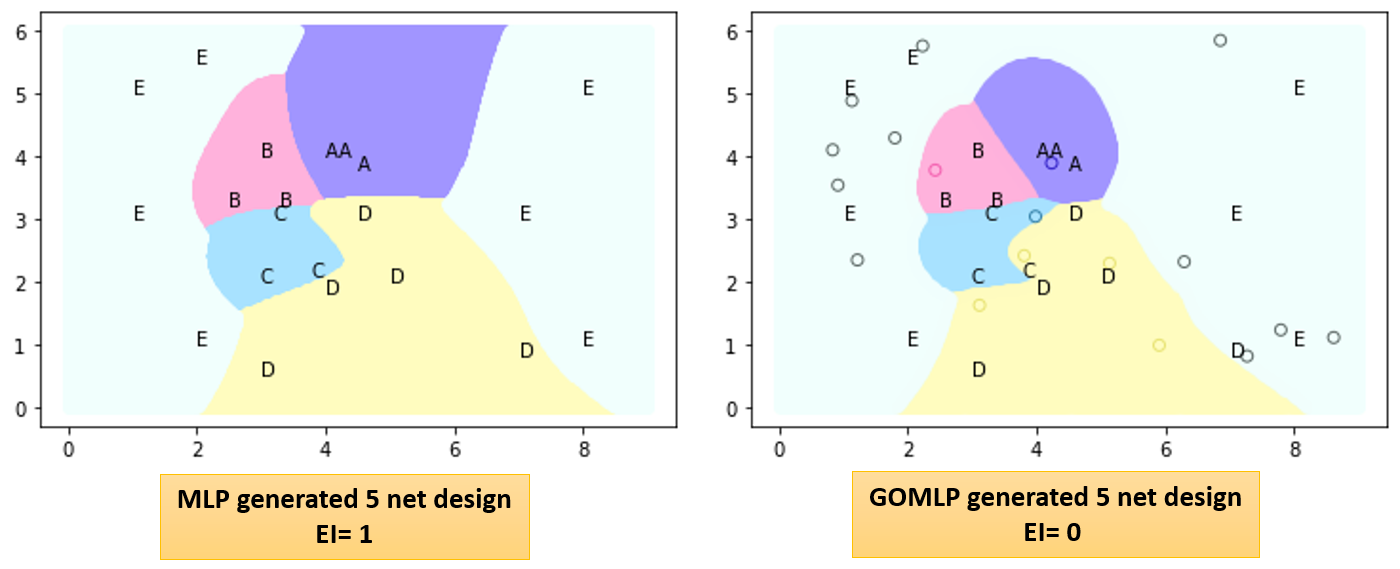}
\caption{Power plane generation on 5 nets PCB layer design case using MLP vs GOMLP. MLP: 6 partitions with 2 disjoint partitions belonging to net E , evolutionary MLP : 5 partitions with no disjoint partitions
}
\label{fig:mlp-mlpga}       
\end{figure}

Fig.~\ref{fig:mlp-mlpga} , shows the results of MLP and GOMLP on a 5 net case. On the left, the partitions are generated using only MLP. Although the MLP can generate a power plane design that correctly classifies the nets. It fails to converge to the most desired power plane partitioning where the partition for each net is connected. As a result, the partition corresponding to net E is divided into two islands. On the right side, the partition is generated using GOMLP. It was executed for just 2 generations and generated the desired result which satisfied constraints. Also, more importantly, the space partitions for each net are all single connected planes. The result demonstrates that the power plane generation is a nontrivial space partition problem that can not be solved directly with a classifier. The challenge for a single classifier to generate a feasible power plane is due to the difficulty of encoding the design objective of 
overall island minimization as a differentiable function into the loss function of MLP. While Euler number~\cite{dyer1980computing} allows the determination of the number of disconnected islands, connecting that loss to the handle parameters is not readily achievable.


In our GOMLP, the combination of MLP and GO solves this challenge by optimizing the handle locations to  control the shape of the MLP generated space partitions. Since handle optimization seeks to minimize the number of islands,  GO will work on connecting isolated islands of the MLP-generated partitions implicitly.  

\subsection{Comparison between GOMLP with and without distance based cost}

\begin{figure}
\centering
\includegraphics[scale=.4]{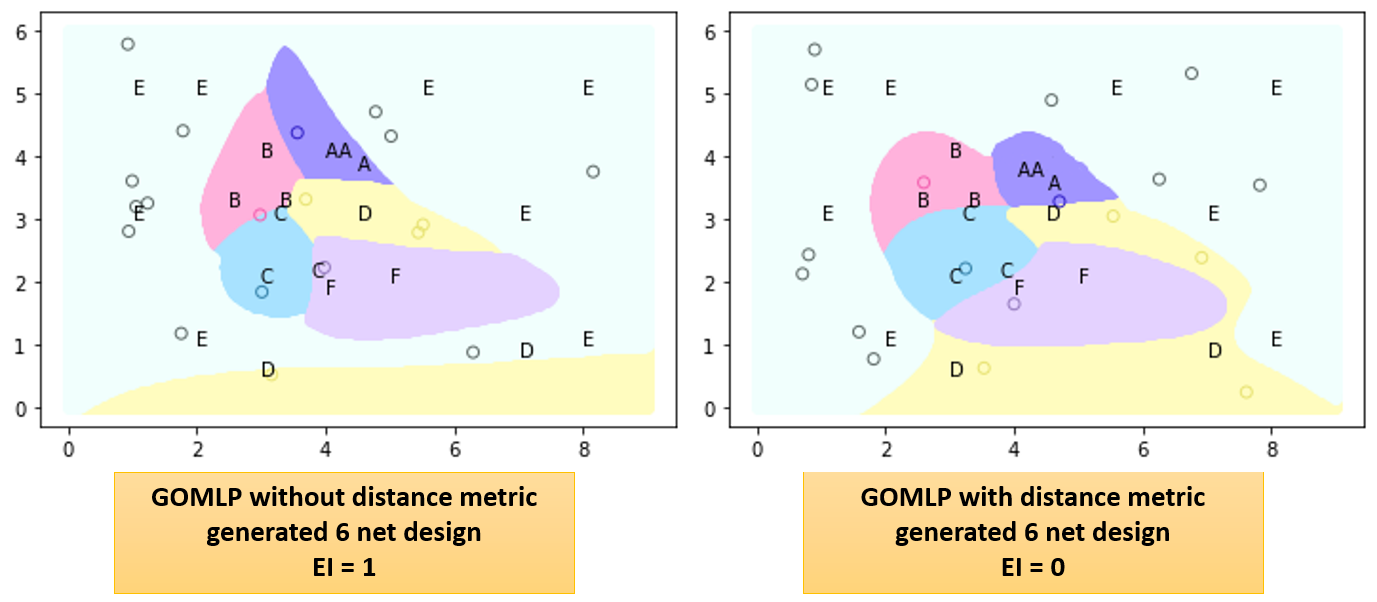}
\caption{Power plane generation on 6 nets PCB layer design using GOMLP with and without distance metric.}
\label{fig:mlpga-mlpgaDM}       
\end{figure}

Fig.~\ref{fig:mlpga-mlpgaDM} shows the experiment results of the GOMLP with and without the distance metric described in the Method part.  A 6-net PCB board layer design is given and the GOMLP is applied twice to solve the problem: the first time with the distance metric and the second time without the distance metric. The distance metric describes the proximity of isolated islands for each net. On the left of Fig.~\ref{fig:mlpga-mlpgaDM}, the generated power plane is based on GOMLP without including distance metric, which means only the number of isolated islands are used to describe the layout of the space partition. On the power plane, the space partition of net E is divided into two isolated islands on the left and right sides of the plane. While on the right of Fig.~\ref{fig:mlpga-mlpgaDM}, the generated power plane is based on GOMLP with the distance metric included. In this case, the space partition corresponding to net E is connected as a single U-shaped plane during the iterative optimization procedure of the GOMLP. 

The improved performance with the introduction of the distance metrics compared to cases where the only number of isolated islands are used in cost function can be explained as follows: in the optimization process of the GOMLP, the space partition is controlled by the layout of the handles. The handles are iteratively optimized to generate higher fitness scores through the GO operation. This mechanism means that the GO is working on optimizing the planes implicitly through the handles' movement. As a result, in cases where the handles for one net are distributed as distant clusters, for instance, the case in Fig.~\ref{fig:mlpga-mlpgaDM}, it would be challenging for the GOMLP to provide the handles with a good moving direction without the introduction of distance metric. In other words, the number of isolated islands is not enough to provide enough guidance for driving the handles to gradually move and form connected space partitions for each net. The distance metric provides something similar to gradient information to guide the handles gradually move in order to connect isolated islands for each net. 

\subsection{Comparison of GOMLP with and without feature expansion}
\begin{figure}
\centering
\includegraphics[scale=.37]{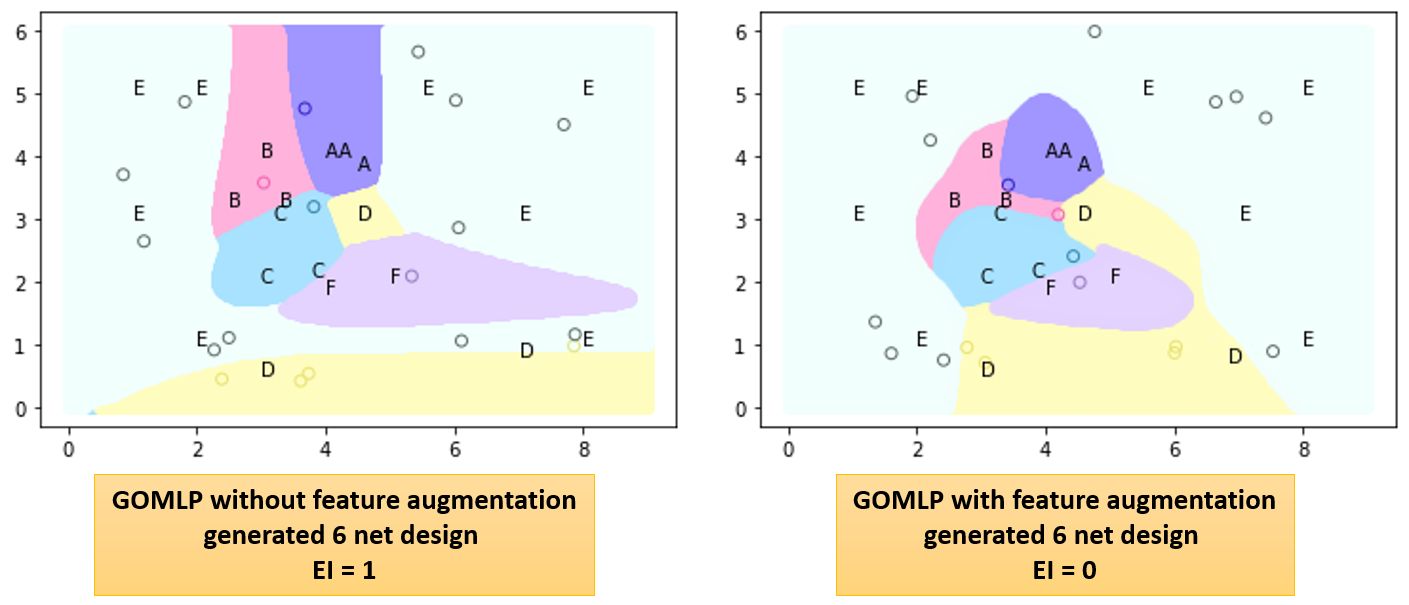}
\caption{Power plane generation on 6 nets PCB layer design using GOMLP with and without feature expansion. GOMLP: 7 partitions with 1 disjoint partitions belonging to net D, with feature expansion: 6 partitions with no disjoint partitions.}
\label{fig:mlpga-mlpgaFE}       
\end{figure}

Fig.~\ref{fig:mlpga-mlpgaFE} shows the results GOMLP with and without feature expansion on a 6 net PCB layer design. On the left, the partitions are generated with GOMLP without the use of feature expansion. Without the expanded features, the model is unable to generate complex space partitions that are feasible and desirable. In the generated power plane, $\Omega_D$ is separated into two isolated islands, which makes it undesirable. Whereas on the right of Fig.~\ref{fig:mlpga-mlpgaFE}, the partitions are generated by GOMLP with feature expansion. With the additional higher order features as input, the MLP can learn more complex space partitions and curved space partitions which leads to desirable power plane generation. In the generated power plane, the partition for  $\Omega_D$ is a singly connected plane with a thin channel connecting the two isolated regions of net D. The comparative results demonstrate the effect of feature expansion in our method. The introduction of high dimensional features makes it easier for the MLP to generate complex space partitions such as zig-zag-shaped planes. 

\subsection{Scalability of GOMLP }
As described in the Experiment part, the second part of the experiments includes the applications of both the  GOMLP and A* method to problems with different configurations and difficulties to systematically assess the performance and scalability of the GOMLP and A*. For the 129 problems experimented, the number of nets ranged from 6 to 8, and problems with more tend nets problems tend to be more difficult. Given a solution, EI is used as the primary metric.  Only when EI equals 0, the generated power plane is desirable. 

\begin{figure}
\centering
\includegraphics[scale=.45]{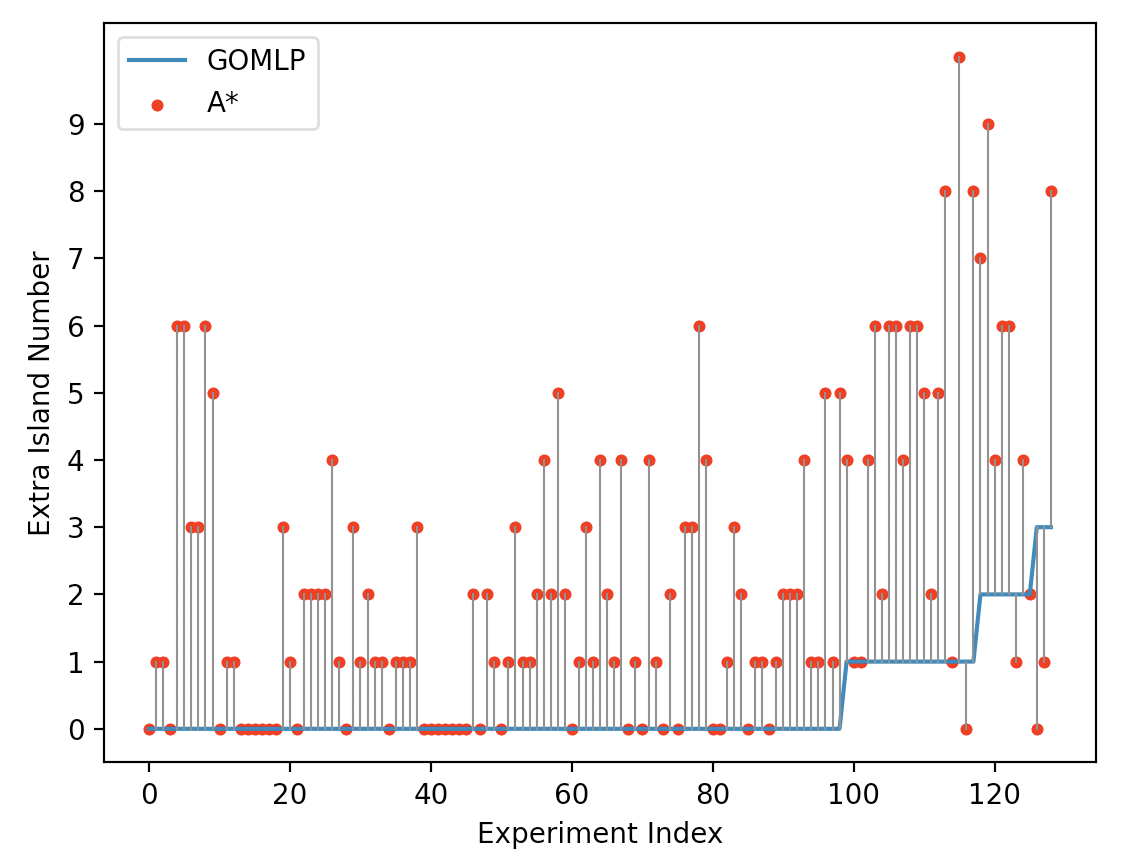}
\caption{GOMLP vs A* for the 129 different board problems. Vertical axis shows the extra islands. GOMLP is better than A* in 71\% of the cases.}
\label{fig:mlp-ga-compareExp}       
\end{figure}

Fig.~\ref{fig:mlp-ga-compareExp} shows the results of GOMLP and A* on the set of 129 problems. The plot shows the number of extra islands for solutions of GOMLP and A* in the ascending order of evolutionary MLP results, as well as their gaps. The plot shows that GOMLP is significantly better in generating high quality power planes with fewer extra islands. Among all the 129 experiments, GOMLP outperforms or has the number of extra islands as A* in 71\% of cases. For problems where A* and GOMLP perform similarly, both methods tend to identify the optimal power planes ($EI=0$). A paired t-test of  GOMLP and A* on our 129 board dataset indicates that GOMLP is statistically better than A* (p<0.001).

A* outperforms GOMLP in only 4 out of the 129 experiments. Detailed studies are conducted on those rare cases where A* outperforms GOMLP. The primary reason that GOMLP did not perform as well as A* in these cases is due to the presence of very near-by pins that belong to different nets, challenging the MLP  to correctly classify these pins. However, these cases will not likely be encountered in realistic PCB board designs given the multilayer flexibility of the problem: designers would move some of the very close pins belonging to different layers of the PCB board to avoid such corner cases. To sum up, GOMLP has good scalability to non-trivial power plane generation designs and performs significantly better than A* baseline.

\begin{figure}
\centering
\includegraphics[scale=.47]{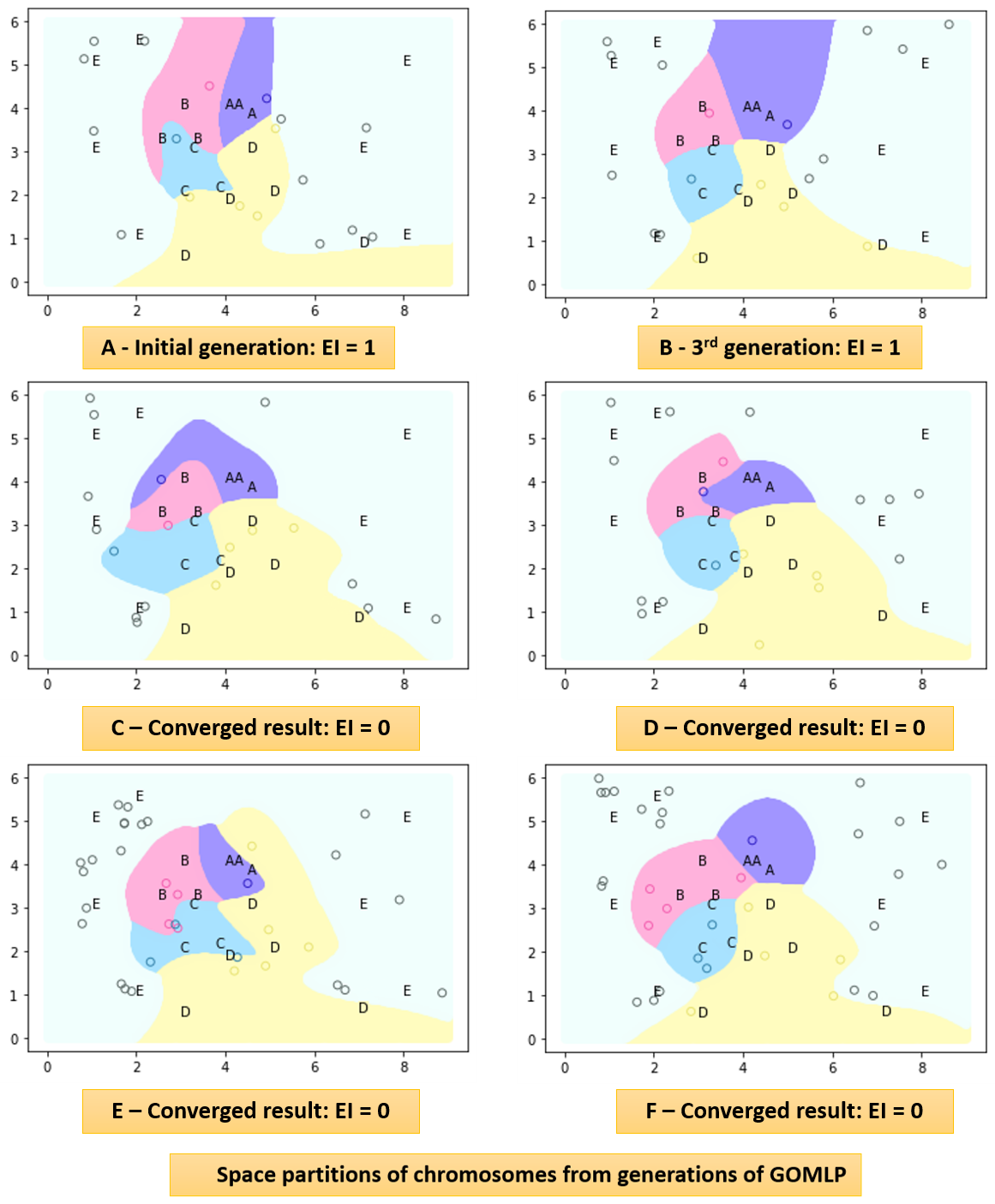}
\caption{Resulting power planes for a 5-net design problem}
\label{fig:mlpga-results-demo}       
\end{figure}

\begin{figure*}[thpb]
\centering
\includegraphics[scale=.5]{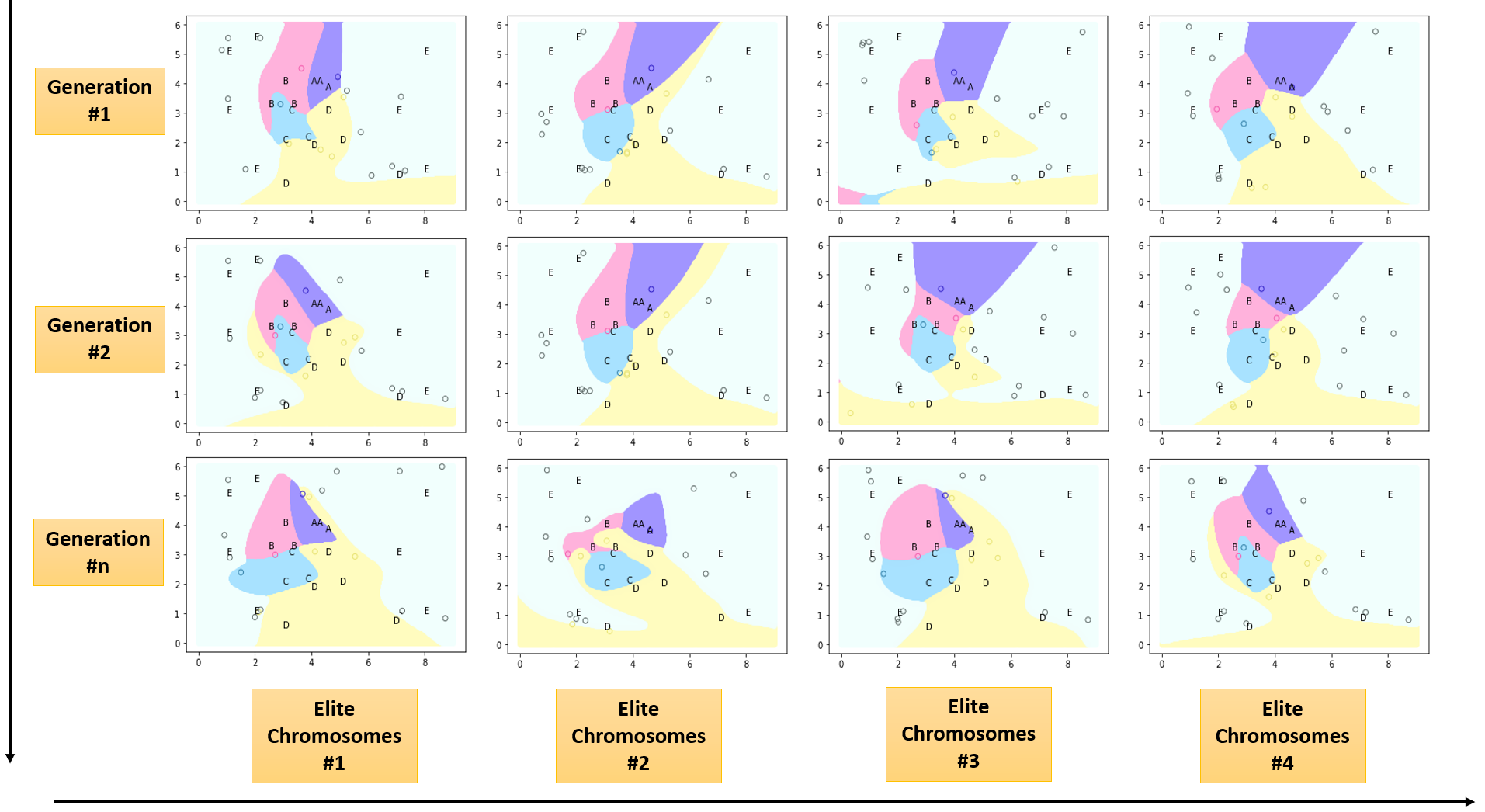}
\caption{Evolution of the handles across the generations. Colored circles: handle locations, letters A-E: pins belonging to different nets. Elite$\#4$ in generation$\#3$ converges to the desired result.}
\label{fig:mlpga-results-dots-evolution}       
\end{figure*}

\begin{figure*}[thpb]
\centering
\includegraphics[scale=.53]{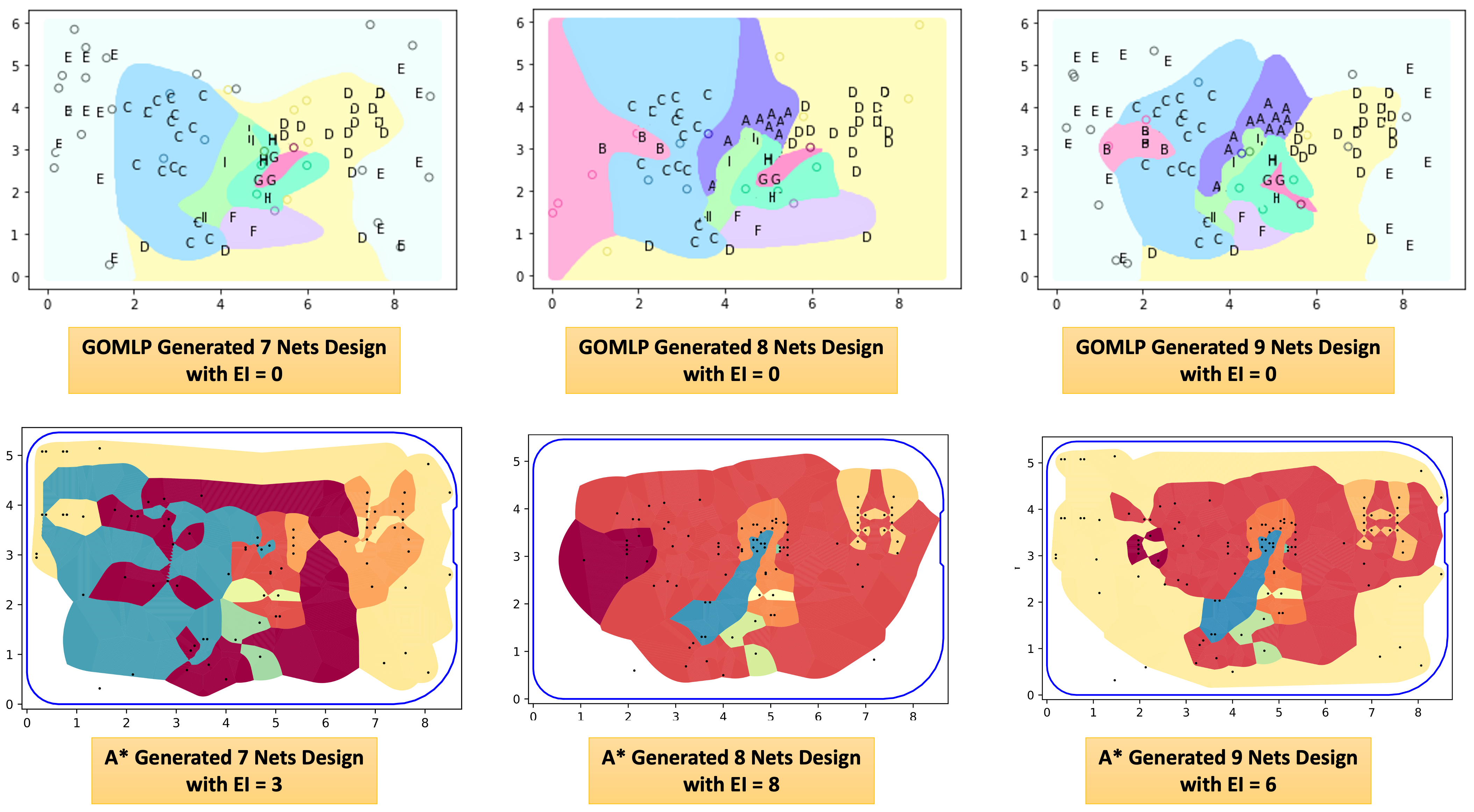}
\caption{GOMLP (top) and A* (bottom) generated power planes for problems with 6,7 and 8 nets.}
\label{fig:mlpga-layout-display}       
\end{figure*}

Finally, to give a more intuitive demonstration of how the GOMLP works, Fig.~\ref{fig:mlpga-results-demo} and Fig.~\ref{fig:mlpga-results-dots-evolution} show the evolution of space partitions and the handles across generations. In Fig.~\ref{fig:mlpga-results-demo}, the pins (annotated with alphabets) are correctly classified in the very first few generations, however, the space partition for net E is divided into two separated islands in the left and right side of the board. As the evolutionary MLP keeps optimizing the space partitions, the two islands for net E start to approach and then finally merged into a desired one single plane. The handles are the implicit driving force for the evolution, as shown in Fig.~\ref{fig:mlpga-results-dots-evolution}, where the handles, pins, and the planes are plotted together. The handles locations changed across generations to implicitly change the space partitions generated by the MLP. It is worth mentioning that the moving patterns of the handles are not as intuitive as we expected. For instance, during the merge of two islands for net E, the handles for net E on two different islands are not directly moving towards the other islands. This is because the relationship between handles and the space partitions is highly nonlinear and it is part of our future work to further study the evolutionary patterns of the handles. For design C,D,E and F in Fig.~\ref{fig:mlpga-results-dots-evolution}, they are all converged solutions for the same problem, showing another advantage for GOMLP: generating multiple solutions for the same problem. Fig.~\ref{fig:mlpga-layout-display} shows the generated power planes by GOMLP and A* on the same set of the problems with 6, 7, and 8 number of nets. For the three problems, GOMLP is able to generate desirable power planes, while A* method generates power planes that have more EI numbers. 

\subsection{\textcolor{black}{H-GOMLP: effectiveness of layer assignment}}
\begin{figure}
\centering
\includegraphics[scale=.49]{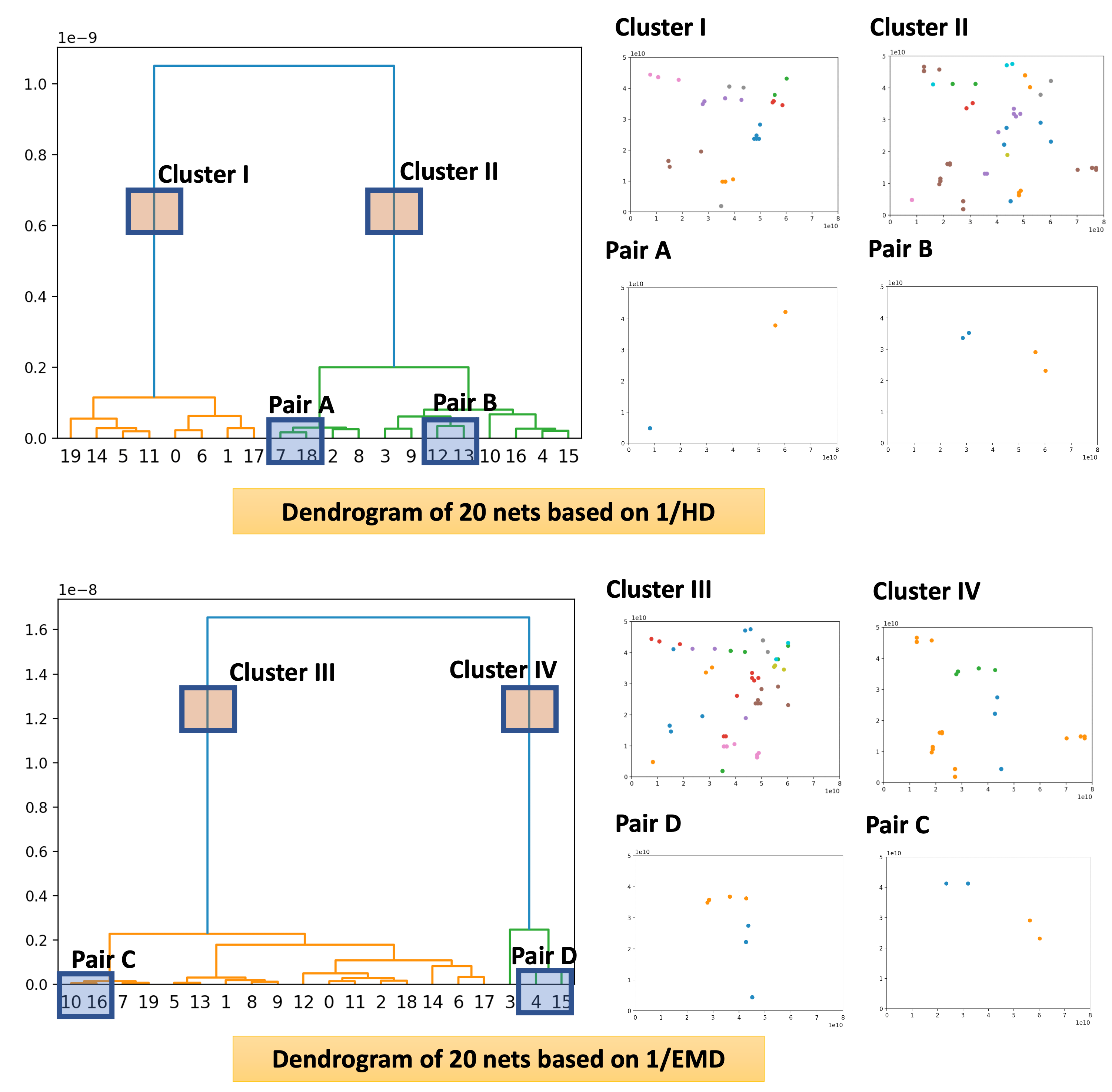}
\caption{Dendrograms from hierarchical clustering of 20 nets in a multilayer power plane generation problem based on 1/HD and 1/EMD. Pair A and Pair C are net pairs that are the first pairs merged; pair B and pair D are net pairs that are the last pairs merged; clusters I,II,III and IV are the clusters of nets that are merged at the highest level on the dendrograms.}
\label{fig:dendro-scatter}       
\end{figure}

\textcolor{black}{To validate the effectiveness of our chosen metrics (EMD and HD) in the hierarchical clustering, we studied the structure of dendrograms generated during hierarchical clustering of the H-GOMLP, especially the sequence by which different pairs of nets are merged on the dendrograms. Fig.~\ref{fig:dendro-scatter} shows dendrograms based on hierarchical clustering of 20 nets in a multilayer power plane generation problem based on the inverse of HD and the inverse of EMD. In the dendrogram based on the inverse of HD, net pair A and net pair B are highlighted and their pins are plotted as scatter plots respectively. Net pair A is the pair that is merged first on the dendrogram and net pair B is the last pair merged. By comparing their pins' scatter plots, it can be seen that pins of net pair A are further away to than those of net pair B, which corresponds to the fact that the HD of net pair A is larger than that of net pair B. In the dendrogram based on the inverse of EMD, the net pair C (highlighted) is the first pair merged on the dendrogram while net pair D (highlighted) is the last pair merged. Their respective scatter plots of pins also show similar trend as those of HD case: pins for net pair C are further away and less overlapping than pins for net pair D. The results show the effectiveness of our chosen metrics in prioritizing the merge of nets with less overlapping area or are further away. Clusters of nets (I,II,III,IV) that are merged at the higher level on the dendrograms are also plotted. 
}

\subsection{\textcolor{black}{H-GOMLP: results on solving non-trivial multilayer problems}}
\begin{figure*}[thpb]
\centering
\includegraphics[scale=.45]{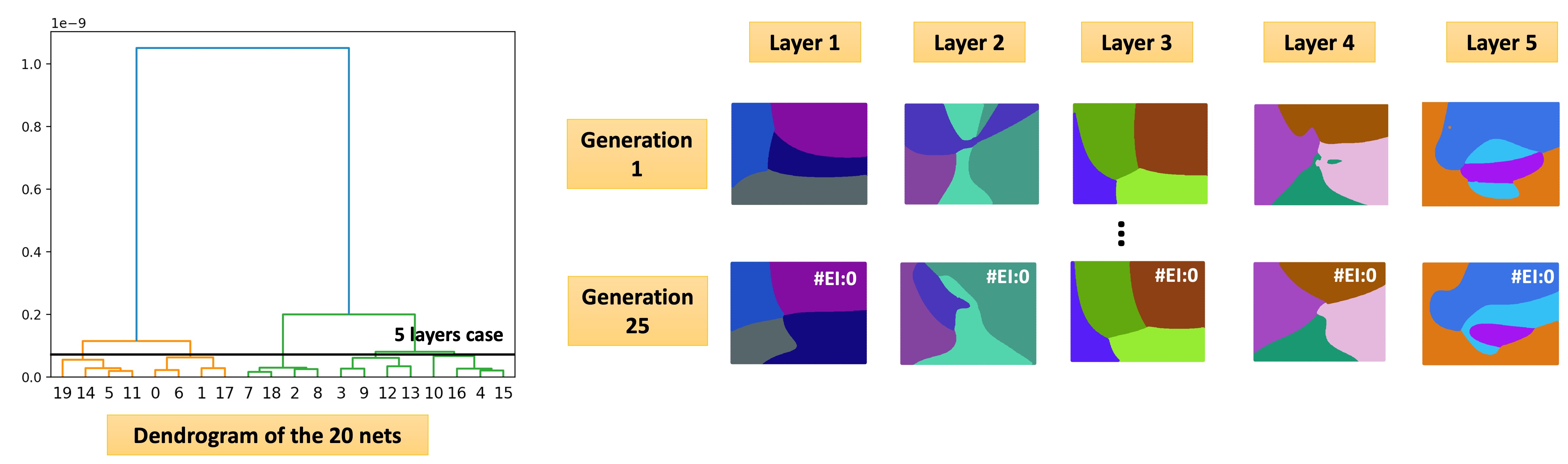}
\caption{H-GOMLP generated dendrogram and multilayer power planes for a 20-nets multilayer board design based on inverse of HD.}
\label{fig:HD-multi-gomlp-20}       
\end{figure*}

\begin{figure*}[thpb]
\centering
\includegraphics[scale=.45]{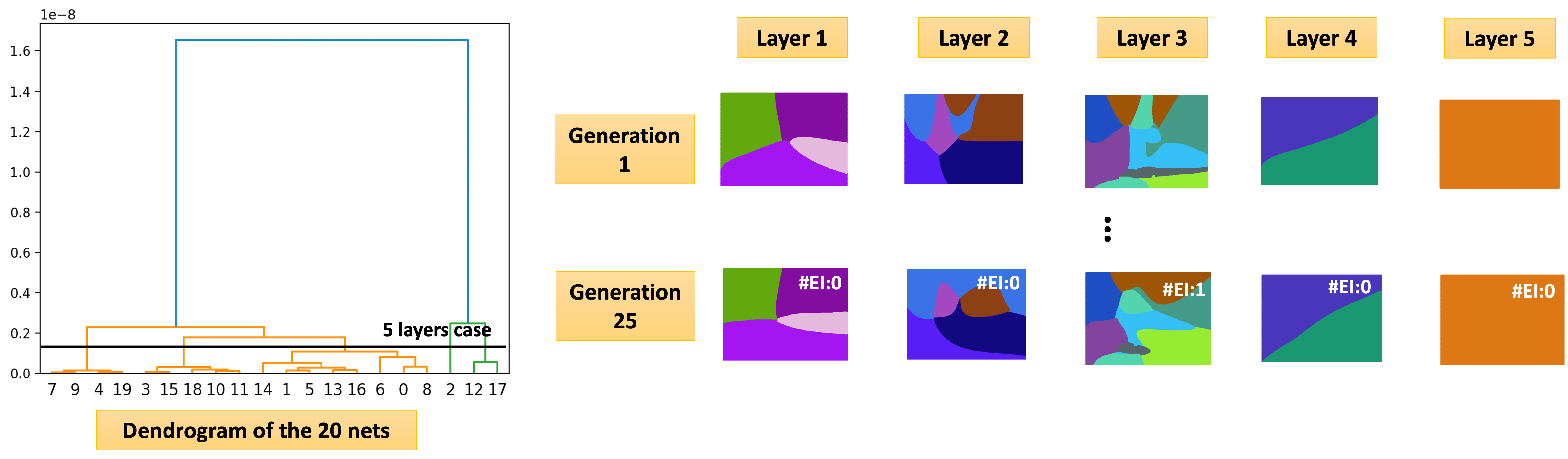}
\caption{H-GOMLP generated dendrogram and multilayer power planes for a 20-nets multilayer board design based on inverse of EMD.}
\label{fig:EMD-multi-gomlp-20}       
\end{figure*}

\textcolor{black}{Fig.~\ref{fig:HD-multi-gomlp-20} shows the results of H-GOMLP on solving a 20-nets multilayer power plane generation problem with HD as the distance metric. The dendrogram is shown on the left and it is marked with the level corresponding to 5 layers. The corresponding power planes for all 5 layers in the first and last generation of GOMLP are shown on the right. With 5 layers, H-GOMLP is able to generate desirable power planes for all 5 layers. We also ran the 4-layers case, and one of the layer ended up having number of EIs larger than 0, which makes the design not desirable. So the MCDL for this problem solved with H-GOMLP (with HD) is 5. For comparison, we also show the results for the same 20-net problem by H-GOMLP based on EMD as distance metric in Fig.~\ref{fig:EMD-multi-gomlp-20}. With the same number of layers, the EMD-based H-GOMLP can achieve desirable designs (EIs=0) for 4 out of the 5 layers while leaving one layer having larger than 0 number of EIs. This indicates that the MCDL for this problem and H-GOMLP (with EMD) is larger than 5. Also, the EMD-based H-GOMLP tends to generate less balanced layer assignment by assigning only one net to Layer 5. The smaller MCDL and more balanced layer assignment of HD-based H-GOMLP both suggest that it is better than the EMD-based H-GOMLP for this problem. }

\subsection{Discussions}

In comparing  GOMLP and A*, one obvious advantage that contributes to the better performance of GOMLP is that it can avoid the combinatorial  challenges that  A* faces: net sequencing, sequence of islands pairs as well as the design of upsampling nodes. It has been shown~\cite{abel1972ordering} that there is no  heuristic for net ordering  that is universally better for all possible problems. This means that methods like A* need to solve the NP-hard combinatorial problem. Even if optimized net sequences or upsamplings are given, the sequential nature of such a method means that the subproblems solved first are prioritized over those solved later, which is not desirable in partitioning problems of this sort~\cite{chen2009global}.  GOMLP provides a good alternative to avoid such combinatorial challenges. Another advantage of the GOMLP over A* is the flexibility to add complex objectives into the method. The A* method is based on generating skeletons of each net using search, which solely focuses on finding the shortest paths, this makes adding more complex objectives such as IR-drop to A* challenging. On the other hand,   GOMLP optimizes over the fitness score in GO, which means the objectives do not have to be explicit and differentiable. This will make sure that the GOMLP can be easily extended to more complex objectives. 

\textcolor{black}{H-GOMLP is able to extend GOMLP to solve more challenging multilayer power plane generation problems. One major advantage of H-GOMLP method comes from its merge-based mechanism. In multilayer PCB board designs, one major challenge is to determine the number of layers needed upfront. In most cases, designers have to rely on tedious trial-and-error approaches which require repeating of the time-consuming multilayer power plane generations with different number of layers, due to lack of proper method to estimate the layer numbers needed. The merge-based nature of H-GOMLP makes the multilayer power plane generatioin a more natural bottom-up procedure, where the number of layers does not need to be determined upfront. Instead, designers can easily get layer assignment choices with incrementally more layers simply by getting clusters from different levels of the dendrogram. It is worth mentioning that for this problem, we don't give the exact optimal power plane generation solution for this 20-nets multilayer power plane generation because of the impractically high computation required. For instance, given a layer number of 5, we need to run GOMLP for each layers of $20^5$ (3.2 million) combinations of layer assignment cases. The optimality gap of H-GOMLP and further improvement of H-GOMLP will be the focus on our future work.}


\section{Conclusion}
In this work, we present a GOMLP automatic power plane generation method based on the combination of MLP and GO. The method is compared against a baseline A* solution. The GOMLP is based on a combination of multilayer perceptron and genetic optimization, with critical components including contour detection, feature expansion as well as the customized distance metric. The GOMLP demonstrates the ability to automatically generate power planes across a diverse set of non-trivial problems with different levels of difficulty. Comparative study also shows the GOMLP is significantly better (in 71\% cases) than A* and those critical components including feature expansion, distance metric as well as the combination of multilayer perceptron and genetic optimization contribute to the competitive capabilities of the GOMLP in automatically generating power planes. \textcolor{black}{We further extend the GOMLP to H-GOMLP which can solve multilayer power plane generation problems by combining GOMLP with hierarchical clustering and customized distance metrics. Future works include adding more complex objectives such as IR-drop to the GOMLP and further improve the performance and scalability of H-GOMLP.}  

\section*{Acknowledgment}
This work is partially funded by the DARPA IDEA program (HR0011-18-3-0010).  
\bibliographystyle{asmems4}  
\bibliography{references}

\end{document}